\documentclass[10pt,twocolumn,letterpaper]{article}

\usepackage{cvpr}              

\usepackage{graphicx}
\usepackage{amsmath}
\usepackage{amssymb}
\usepackage{booktabs}

\usepackage[numbers,sort]{natbib}
\usepackage{graphicx}
\usepackage{mathtools}
\usepackage{tabularx}
\usepackage{multirow}
\usepackage{enumitem}
\usepackage{bbm}
\usepackage{xcolor, colortbl}
\usepackage{wrapfig}
\usepackage{adjustbox}
\usepackage[accsupp]{axessibility}  

\usepackage{subfiles}
\usepackage{afterpage}

\usepackage{algorithm}
\usepackage{algpseudocode}

\usepackage[pagebackref,breaklinks,colorlinks]{hyperref}

\definecolor{grey}{rgb}{0.9, 0.9, 0.9}
\newcommand{\ccol}{\cellcolor{grey}}
\usepackage{pifont}         

\def\ie{\emph{i.e.}}
\def\eg{\emph{e.g.}}
\def\etal{\emph{et al.}}

\def\st{\emph{s.t.}}

\definecolor{grn}{rgb}{0.1, 0.6, 0.1}
\definecolor{mgt}{rgb}{0.8, 0.1, 0.8}

\newcommand{\suhac}[1]{{\color{grn}{(#1)}}}

\newcommand{\sy}[1]{{\color{purple}{#1}}}

\definecolor{bittersweet}{rgb}{1.0, 0.44, 0.37}

\definecolor{new_blue}{rgb}{0.14, 0.47, 0.74}
\definecolor{new_pink}{rgb}{1, 0.337, 0.337}
\definecolor{new_lightgreen}{rgb}{0.475, 0.667, 0.25}
\makeatletter
\newcommand{\multiline}[1]{%
  \begin{tabularx}{\dimexpr\linewidth-\ALG@thistlm}[t]{@{}X@{}}
    #1
  \end{tabularx}
}
\makeatother

\usepackage[capitalize]{cleveref}
\crefname{section}{Sec.}{Secs.}
\Crefname{section}{Section}{Sections}
\Crefname{table}{Table}{Tables}
\crefname{table}{Tab.}{Tabs.}

\def\cvprPaperID{3831} 
\def\confName{CVPR}
\def\confYear{2022}

\begin{document}

\title{Self-Taught Metric Learning without Labels}

\author{
Sungyeon Kim$^1$ \qquad
Dongwon Kim$^1$ \qquad
Minsu Cho$^{1,2}$ \qquad
Suha Kwak$^{1,2}$ \\
Dept. of CSE, POSTECH$^1$ \ \ \ \ \ \ \ \ \ \ \ \ \ \ Graduate School of AI, POSTECH$^2$ \\
{\tt\small \{sungyeon.kim, kdwon, mscho, suha.kwak\}@postech.ac.kr} \\
{\tt\small \url{http://cvlab.postech.ac.kr/research/STML/}}
}
\maketitle


\begin{abstract}

We present a novel self-taught framework for unsupervised metric learning, which alternates between predicting class-equivalence relations between data through a moving average of an embedding model and learning the model with the predicted relations as pseudo labels.
At the heart of our framework lies an algorithm that investigates contexts of data on the embedding space to predict their class-equivalence relations as pseudo labels.
The algorithm enables efficient end-to-end training since it demands no off-the-shelf module for pseudo labeling. 
Also, the class-equivalence relations provide rich supervisory signals for learning an embedding space.
On standard benchmarks for metric learning, it clearly outperforms existing unsupervised learning methods and sometimes even beats supervised learning models using the same backbone network.
It is also applied to semi-supervised metric learning as a way of exploiting additional unlabeled data, and achieves the state of the art by boosting performance of supervised learning substantially.
\end{abstract}


\section{Introduction}
\label{sec:intro}



Understanding similarities between data is at the heart of many machine learning tasks such as data retrieval~\cite{kim2019deep,movshovitz2017no,Sohn_nips2016,songCVPR16}, face verification~\cite{liu2017sphereface,Schroff2015}, person re-identification~\cite{Chen_2017_CVPR,xiao2017joint}, few-shot learning~\cite{Qiao_2019_ICCV, snell2017prototypical, sung2018learning}, and representation learning~\cite{kim2019deep,Wang2015,Zagoruyko_CVPR_2015}.
Metric learning embodies the perception of similarity by learning an embedding space where the distance between a pair of data represents their inverse semantic similarity.
Also, the mapping from data to such a space is typically modeled by deep neural networks. 

Recent advances in metric learning rely heavily on supervised learning using large-scale datasets.
Manual annotation of such datasets is however costly, and thus could limit the class diversity of training data and in consequence the generalization capability of learned models. 
Unsupervised metric learning has been studied to resolve this issue.
Existing methods in this line of research mainly synthesize class information of training data by assigning a surrogate class per training instance~\cite{dosovitskiy2015discriminative, wu2018unsupervised, ye2019unsupervised, ye2020probabilistic} or discovering pseudo classes through $k$-means clustering~\cite{cao2019unsupervised, caron2018deep, li2020unsupervised, yan2021unsupervised, kan2021relative, li2021spatial}, hierarchical clustering~\cite{yan2021unsupervised}, or random walk~\cite{iscen2018mining}.
Although these methods have demonstrated impressive results without using groundtruth labels in training, they often fail to
capture intra-class  variation~\cite{dosovitskiy2015discriminative, wu2018unsupervised, ye2019unsupervised, ye2020probabilistic} or impose substantial computational burden due to the off-the-shelf techniques~\cite{cao2019unsupervised, caron2018deep, li2020unsupervised, yan2021unsupervised, kan2021relative, iscen2018mining, li2021spatial}.

\begin{figure} [!t]
\centering
\includegraphics[width = 0.95\columnwidth]{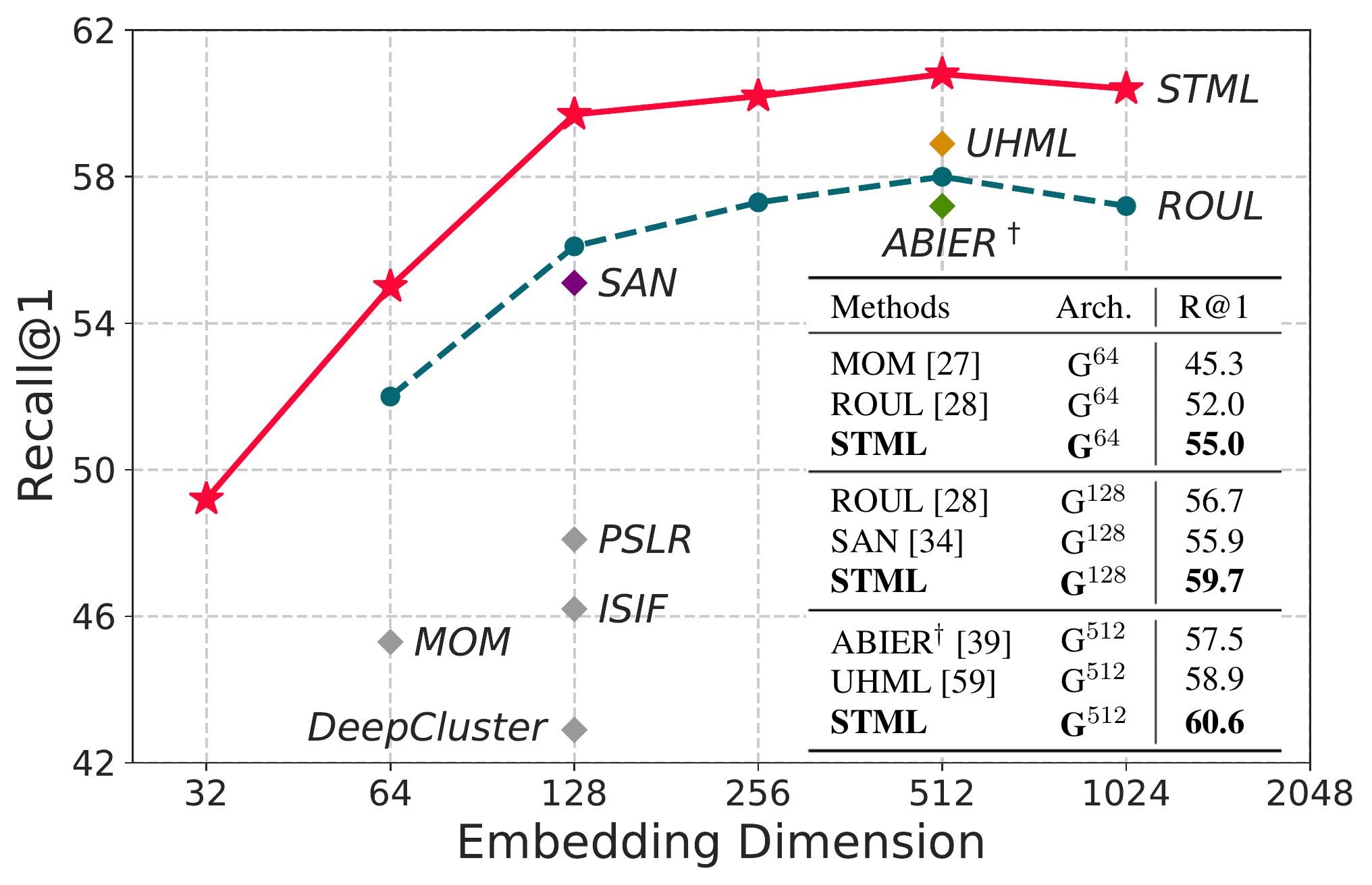}
\vspace{-2mm}
\caption{Accuracy in Recall@1 versus embedding dimension on the CUB-200-2011~\cite{CUB200} dataset using GoogleNet~\cite{Szegedy2015} backbone. Superscripts denote embedding dimensions and $\dagger$ indicates supervised learning methods. Our model with 128 embedding dimensions outperforms all previous arts using higher embedding dimensions and sometimes surpasses supervised learning methods.
}
\label{fig:teaser}
\vspace{-2mm}
\end{figure}

\begin{figure*} [!t]
\centering
\includegraphics[width = 0.99 \textwidth]{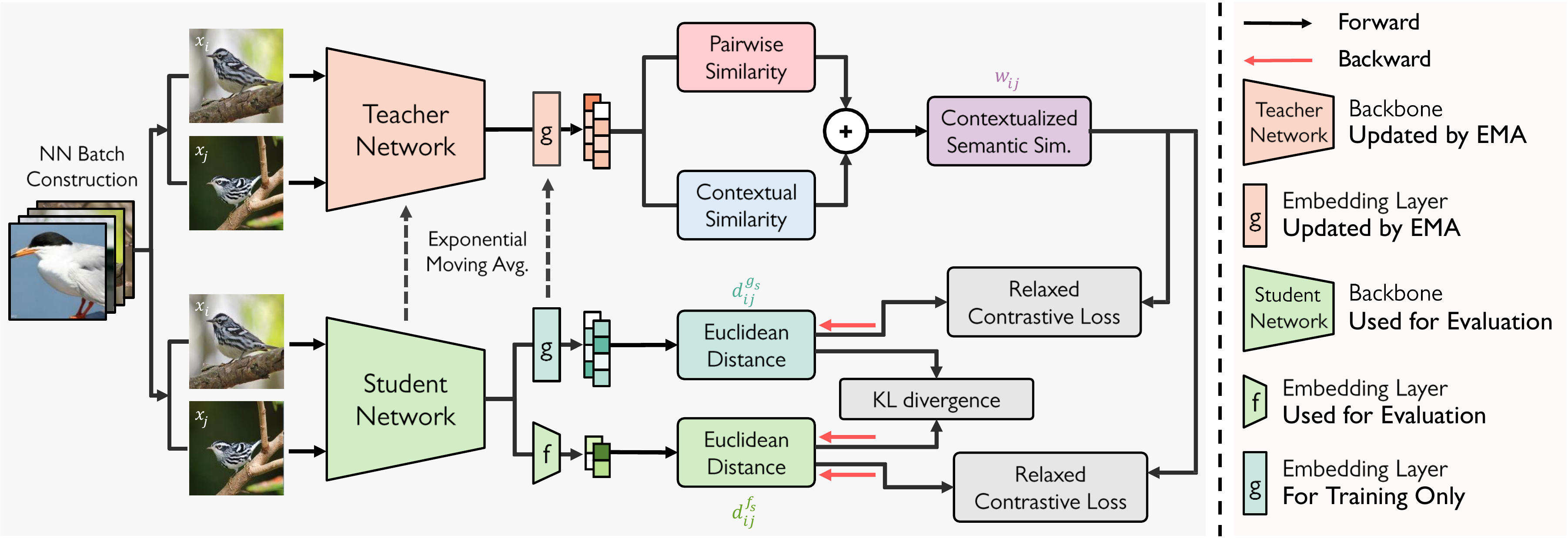}
\vspace{-1mm}
\caption{An overview of our STML framework. 
First, contextualized semantic similarity between a pair of data is estimated on the embedding space of the teacher network. The semantic similarity is then used as a pseudo label, and the student network is optimized by relaxed contrastive loss with KL divergence. {\color{new_pink}Pink} arrows represent backward gradient flows. Finally, the teacher network is updated by an exponential moving average of the student. The student network learns by iterating these steps a number of times, and its backbone and embedding layer in {\color{new_lightgreen}light green} are considered as our final model.
} 
\label{fig:our_framework}
\vspace{-2mm}
\end{figure*}

In this paper, we propose a new method for unsupervised metric learning that addresses the aforementioned limitations of previous work and achieves the state of the art as shown in Fig.~\ref{fig:teaser}.
The major contribution of this paper is two-fold.
First, we introduce \emph{a novel, end-to-end self-taught metric learning framework} (STML), whose overall pipeline is illustrated in Fig.~\ref{fig:our_framework}.
Unlike the existing works, it predicts class-equivalence relations between data within each mini-batch by \emph{self-exploration} and leverages the predicted relations as synthetic supervision for metric learning.
Our framework manages two embedding networks, namely teacher and student whose backbone are initialized identically.
Class-equivalence between a pair of data is approximately estimated as their semantic similarity on the embedding space of the teacher network.
The predicted similarity is used as a soft pseudo label for learning the student model, and the teacher model is in turn updated by a momentum-based moving average of the student.
Iterating this process evolves the student model progressively.

The success of STML depends heavily on the quality of predicted semantic similarities, and our second contribution lies in the way of \emph{estimating semantic similarities using contexts}.
Specifically, given a pair of data, we compute their semantic similarity considering the overlap of their contexts (\ie, neighborhoods in the embedding space) as well as their pairwise distance.
We found that the contextualized semantic similarity approximates class-equivalence precisely.
Further, since it has a real value indicating the degree of semantic similarity, it provides rich information beyond binary class-equivalence.
Also, since it requires no external module nor memory banks~\cite{he2020momentum, koohpayegani2021mean}, it makes STML efficient and concise.
To further enhance the quality of predicted semantic similarity, we design the teacher network to learn a higher dimensional embedding space than the student counterpart.
This asymmetric design of the two networks allows the teacher to provide more effective supervision thanks to its improved expression power while the student, \ie, our final model, remains compact.

To the best of our knowledge, STML is the only unsupervised metric learning method that can consider semantic relations between data in end-to-end training without introducing off-the-shelf techniques.
Compared to the instance-level surrogate classes~\cite{dosovitskiy2015discriminative, wu2018unsupervised, ye2019unsupervised, ye2020probabilistic}, pseudo labels generated and exploited by STML are more appropriate to capture semantic relations between data since they indicate class-equivalence relations.
Also, unlike previous work based on pseudo labeling~\cite{cao2019unsupervised, caron2018deep, iscen2018mining,li2020unsupervised,  yan2021unsupervised, kan2021relative}, STML employs no external algorithm and thus allows training to be efficient, end-to-end, and less sensitive to hyper-parameters.
Further, it is naturally applied to semi-supervised metric learning~\cite{duan2020slade} as well as unsupervised metric learning. 





We first evaluate STML on standard benchmarks for metric learning~\cite{krause20133d,songCVPR16,CUB200}, where it largely outperforms existing unsupervised learning methods.
Surprisingly, sometimes it even beats some of supervised learning models using the same backbone network as shown in Fig.~\ref{fig:teaser}. 
Beyond that, its efficacy is demonstrated on two benchmarks for semi-supervised metric learning~\cite{duan2020slade}, where it substantially outperforms previous work as well.


\section{Related Work}
\label{sec:relatedwork}

\noindent \textbf{Unsupervised metric learning}
has been mainly addressed in two different directions, instance discrimination~\cite{dosovitskiy2015discriminative, wu2018unsupervised, ye2019unsupervised, ye2020probabilistic} and pseudo labeling~\cite{cao2019unsupervised, caron2018deep, li2020unsupervised, yan2021unsupervised, kan2021relative, iscen2018mining, li2021spatial}.
Following the contrastive learning strategy~\cite{bachman2019learning, chen2020simple,chen2020improved,he2020momentum}, instance discrimination methods assign a unique label per training instance and learn embedding spaces where different instances are well discriminated. 
Unfortunately, they have a trouble modeling variations within each latent class.
On the other hand, pseudo labeling methods discover pseudo classes by applying off-the-shelf algorithms like $k$-means clustering~\cite{cao2019unsupervised, caron2018deep,li2020unsupervised, li2021spatial, kan2021relative}, hierarchical clustering~\cite{yan2021unsupervised} and random walk~\cite{iscen2018mining} to unlabeled training data.
Such methods can consider the class-equivalence relations between training data by utilizing existing supervised metric learning losses~\cite{Hadsell2006, Schroff2015, wang2019multi} with pseudo labels.
However, the major drawback of these methods is their prohibitively high complexity imposed by the auxiliary algorithms, which obstructs end-to-end training.

\noindent \textbf{Semi-supervised metric learning}, 
recently introduced by Duan~\etal~\cite{duan2020slade}, assumes that a training dataset consists of both labeled and unlabeled data.
The previous work discovers and assigns pseudo labels to unlabeled training data via $k$-means clustering. Unfortunately, it also suffers from the aforementioned drawbacks of the unsupervised learning using pseudo labeling.

\noindent \textbf{Self-supervised representation learning}
aims at learning high-quality intermediate features to generalize to other downstream tasks.
To this end, various annotation-free pretext tasks have been proposed, \eg, rotation prediction~\cite{gidaris2018unsupervised}, solving jigsaw puzzle~\cite{noroozi2016unsupervised}, image colorization~\cite{zhang2016colorful}, and image inpainting~\cite{pathak2016context}. 
Pseudo labeling also has been widely used as it allows to learn models using conventional losses; existing methods in this direction discover pseudo classes by clustering~\cite{caron2018deep} or treat each training instance as a surrogate class~\cite{chen2020simple,dosovitskiy2015discriminative,he2020momentum, wu2018unsupervised}.
Also, as an extension of the surrogate class, nearest neighbors have been leveraged as a positive set for self-supervised learning~\cite{huang2019unsupervised, dwibedi2021little, koohpayegani2021mean}, but they often produce noisy labels unless a sufficiently trained embedding space is provided in advance.

\section{Proposed Method}
\label{sec:method}

This section first presents an overview of our self-taught metric learning framework, dubbed STML. We then describe the each step of STML in detail and how to estimate the contextualized semantic similarity, which is the synthetic supervision for our framework.

\subsection{Overview}

STML 
enables efficient end-to-end training without using any off-the-shelf technique, while capturing semantic relations between data by generating and exploiting pseudo labels that approximate their class-equivalence. 

Fig.~\ref{fig:our_framework} illustrates the overall pipeline of STML. 
It manages two embedding networks with the identically initialized backbone: a \emph{teacher} and a \emph{student}. The teacher network is used to estimate semantic similarities between data as their approximate class-equivalence relations, which are in turn used as synthetic supervision for training the student network. 
In particular, the teacher model leverages a high-dimensional (\eg, 1024 or 2048) embedding layer $g^t$ with the backbone encoder $\phi^t$ of the teacher. The use of the high dimensional space allows the teacher embedding layer to encode richer information and produce more reliable semantic similarity. Our student model has two parallel embedding layers $f^s$ and $g^s$ that share the backbone encoder $\phi^s$. $f^s$ is the embedding layer of our final model, and has a lower dimension (\eg, 128 or 512) to learn a compact embedding space. $g^s$ is an auxiliary layer dedicated to continually updating the teacher embedding layer $g^t$, and thus having the same output dimension with that of $g^t$.


Algorithm~\ref{algo:stml} describes the procedure of STML in detail. First, the teacher network estimates semantic similarities that approximate class-equivalence relations between data, which are in turn used as synthetic supervision for training the student network.
Then the teacher network is updated by the momentum-based moving average of the student.
Our final model is the student network obtained by iterating this alternating optimization a predefined number of times.
In the next section, we will elaborate each step of STML.

\begin{algorithm}[!t]
\fontsize{8.5}{10.5}\selectfont
\linespread{2}
\renewcommand{\algorithmicrequire}{\textbf{Input:}}
\renewcommand{\algorithmicensure}{\textbf{Output:}}
\algrenewcommand\algorithmicindent{1.15em}%

\caption{Self-taught metric learning}
\label{algo:stml}
\begin{algorithmic}[1]
\Require \multiline{teacher model $t$, student model $s$, kernel bandwidth $\sigma$, \\ momentum $m$, batch size $n$.}
\State \textbf{set} $\theta_{t} = \theta_{\phi^{t}} \cup \theta_{g^{t}}, \theta_{s} = \theta_{\phi^{s}} \cup \theta_{g^{s}}$
\State Identically initialize the backbone of $t$ and $s$.
\For{$\emph{number of epochs}$}
    \State \multiline{Construct mini-batches using \emph{nearest neighbor} \\ \emph{batch construction}. \Comment{Step \#1}}
    \For{$\emph{number of iterations}$}
        \For {\textbf{all} samples in a mini-batch $\{x_k\}_{k=1}^n$}
        \State $z^t_k \,\, \leftarrow (g^t \circ \phi^t)(x_k)$  
        \State $z^{f_s}_k \leftarrow (f^s \circ \phi^s)(x_k)$, $z^{g_s}_k \leftarrow (g^s \circ \phi^s)(x_k)$
        \EndFor
        \For{\textbf{all} $i \in \{1, \cdots, n\}$ and $j \in \{1, \cdots, n\}$}
        \Comment{Step \#2}
        \State $w_{ij}^P \leftarrow \exp(-{{||z_i^t - z_j^t||_2^2}}/{\sigma})$
        \State \textbf{compute} $w_{ij}^C$ using Eq.~(\ref{eq:Jaccard_weight}, \ref{eq:Jaccard_asymmetric_weight}, \ref{eq:Jaccard_symmetric_weight}).
        \State $w_{ij} \leftarrow \frac{1}{2}(w^{P}_{ij} + w^{C}_{ij})$
        \EndFor
        \State \textbf{compute} $\mathcal{L}_\textrm{STML}$ using Eq.~\eqref{eq:STML_loss} and \textbf{optimize} $t$.
        \Comment{Step \#3}
        \State $\theta_{t} \leftarrow{} m \theta_{t} + (1-m)\theta_{s}$
        \Comment{Step \#4}
    \EndFor
\EndFor
\State \Return  student model $s$
\end{algorithmic}
\end{algorithm}
\setlength{\textfloatsep}{10pt}

\subsection{Self-Taught Metric Learning Framework}
\label{subsec:self_taught_metric_learning}

\noindent \textbf{Step \#1: Nearest neighbor batch construction.}
The way of constructing mini-batches for training has a massive impact on performance in metric learning~\cite{Ge2018DeepML,roth2020revisiting}.
%
One of the most widely used methods for batch construction is the $pk$-sampling: It randomly chooses $p$ unique classes, and then randomly samples $k$ instances per class. Since this method leverages class labels of training data, it is however prohibited in an unsupervised learning setting. We instead propose a new strategy based on the nearest neighbor search. 
For each mini-batch, it randomly samples $q$ queries and then searches for $k-1$ nearest neighbors of each query; including both queries and their nearest neighbors, the mini-batch becomes of size $qk$. Our strategy is simple, fast, yet allows mini-batches to include diverse and relevant samples.

\noindent \textbf{Step \#2: Generating synthetic supervision.}
Supervised metric learning methods in general employ as supervision class-equivalence relations between data, which are however not available in our setting.
STML instead approximates the class-equivalence relations by semantic similarities between data on the teacher embedding space, which are used as synthetic soft supervision for learning the student model. 
The major challenge is that the teacher embedding space is not mature enough to grasp the semantic relations between data at early stages of unsupervised learning.
We thus propose contextualized semantic similarity, which measures the similarity between a pair of data while considering their contexts on the teacher embedding space;
its details are presented in Sec.~\ref{subsec:similarity}.

\noindent \textbf{Step \#3: Learning the student model.}
We simultaneously optimize the two branches of the student model using the contextualized semantic similarities as synthetic supervision. Since its value is soft, however, the contextualized semantic similarity cannot be incorporated with the ordinary metric learning losses that take discrete labels as supervision.
We thus employ the relaxed contrastive loss~\cite{kim2021embedding}, which is designed to utilize soft relational labels for metric learning. 
Let $w_{ij}$ be the contextualized semantic similarity between $x_i$ and $x_j$, and $z_i^{f_s}$ denote the student embedding vector of $x_i$ generated by $f^s$. Defining $d_{ij}^{f_s}:={||z_i^{f_s} - z_j^{f_s}||_2}/({\frac{1}{n}\sum_{k=1}^{n}||z_i^{f_s} - z_k^{f_s}||_2})$ as the relative distance between $f_i^s$ and $f_j^s$, the loss is then given by
\begin{equation}
\begin{split}
\mathcal{L}_\textrm{RC}(\mathcal{Z}^{f_s}) &= \frac{1}{n}\sum_{i=1}^n\sum_{j\neq i}^n{{w_{ij} \big(d_{ij}^{f_s}}\big)^2} \\
&+ \frac{1}{n}\sum_{i=1}^n\sum_{j\neq i}^n{{(1-w_{ij})\big[\delta - d_{ij}^{f_s}}\big]_+^2},
\end{split}
\label{eq:smooth_contrastive_mu}
\end{equation}
where $\mathcal{Z}^{f_s}$ is all student embedding vectors generated by $f^s$, $n$ is the number of samples in the batch, and $\delta$ is a margin.
The contextualized semantic similarity serves as the weights for the attracting and repelling terms of the loss, and thus determines the magnitude of the force that pulls or pushes a pair of data on the student embedding space.
Note that $g_s$ is trained in the same manner on $f_s$.
We additionally apply self-distillation to further utilize complex higher dimensional information for lower dimensional space. Following \cite{roth2021simultaneous}, we use Kullback-Leibler divergence as objective of self-distillation:
\begin{equation}
\begin{split}
\mathcal{L}_\textrm{KL}(\mathcal{Z}^{f_s}, \mathcal{Z}^{g_s}) &=
\frac{1}{n}\sum_{i=1}^n\sum_{j\neq i}^n\psi(-d_{ij}^{g_s})\log{\frac{\psi(-d_{ij}^{g_s})}{\psi(-d_{ij}^{f_s})}},
\end{split}
\label{eq:KL_divergence}
\end{equation}
where $\psi(\cdot)$ is softmax operation. Note that the gradient-flow of $g_s$ is truncated. In summary, STML trains the two branches $f_s$ and $g_s$ of the student model by minimizing the overall loss defined as
\begin{equation}
\begin{split}
\mathcal{L}_\textrm{STML}(\mathcal{Z}^{f_s}, \mathcal{Z}^{g_s}) &=
\frac{1}{2}\big[
\mathcal{L}_\textrm{RC}(\mathcal{Z}^{f_s}) + \mathcal{L}_\textrm{RC}(\mathcal{Z}^{g_s})\big] \\
&+ \mathcal{L}_\textrm{KL}(\mathcal{Z}^{f_s}, \mathcal{Z}^{g_s}).
\end{split}
\label{eq:STML_loss}
\end{equation}

\noindent \textbf{Step \#4: Momentum update of the teacher model.}
The teacher model should be updated as the student model is trained so that the synthetic supervision computed from the teacher embedding space is enhanced progressively.
The simplest update strategy is to replace the teacher model with the student counterpart at every iteration.
However, this could deteriorate the consistency of the synthetic supervision over iterations, which leads to unstable training of the student model.
We thus update the teacher model by the momentum-based moving average of the student, except for the embedding layer $f^s$ since the teacher model has only a high-dimensional one. Let $\theta_t$ be the parameters of the teacher model and $\theta_t$ be the parameters of $\phi_t$ and $g_t$. Then $\theta_t$ is updated by
\begin{equation}
\theta_t \xleftarrow{} m \theta_t + (1-m)\theta_s,
\label{eq:teacher_momentum_update}
\end{equation}
where $m \in [0,1]$ is the coefficient that controls the momentum update rate, slowing down the update of the teacher model as its value increases.

\subsection{Contextualized Semantic Similarity}
\label{subsec:similarity}

The quality of predicted semantic similarities is of vital importance in STML since they are the only supervision for training.
However, it is not straightforward to compute reliable semantic similarities especially at early stages of unsupervised learning where the teacher network is significantly incomplete.
To overcome this challenge, we propose to use \emph{context}, \ie, neighborhood on a data manifold.

Our key idea is not to rely on the relations considering two images of interest only, but to figure out their indirect semantic relations via their contexts. Their multiple indirect relations can contribute to capturing common patterns frequently observed within each class and thus understanding the underlying class-equivalence relations. To this end, we design the \emph{contextualized semantic similarity}. Given a pair of samples, their contextualized semantic similarity is defined as a combination of their \emph{pairwise similarity} and \emph{contextual similarity} on the teacher embedding space.

\noindent \textbf{Pairwise similarity.}
The pairwise similarity of two samples $x_i$ and $x_j$, denoted by $w_{ij}^P$, is given by
\begin{equation}
w_{ij}^P = \exp\bigg(-\frac{{||z_i^t - z_j^t||_2^2}}{\sigma}\bigg),
\label{eq:pairwise_similarity}
\end{equation}
where $z_i^t$ is the teacher embedding vector of $x_i$ and $\sigma$ is the Gaussian kernel bandwidth. 
This kind of similarity has been known to encode rich information about relations of $x_i$ and $x_j$ when the teacher embedding space is pre-trained in a fully supervised manner~\cite{kim2021embedding}, but is often broken in unsupervised metric learning.
We thus supplement it with the contextual similarity, which is detailed below.

\noindent \textbf{Contextual similarity.}
The underlying assumption of the contextual similarity is that the more two samples semantically similar, the larger their contexts overlap.
One may consider nearest neighbors of a data point on the teacher embedding space as its context, but they often include irrelevant data when the embedding space is not sufficiently trained yet.
We thus instead consider $k$-reciprocal nearest neighbors as context since they are known to discover highly correlated candidates~\cite{qin2011hello,zhong2017re}.
The set of $k$-reciprocal nearest neighbors of a data point $x_i$ is given by
\begin{equation}
R_k(i) = \{j|(j \in N_k(i))\wedge (i \in N_k(j))\},
\label{eq:Reciprocal_knn}
\end{equation}
where $N_k(i)$ is  the $k$-nearest neighbors of $x_i$, including itself (\ie, $i\in N_k(i)$), in the current mini-batch.
%
The initial form of contextual similarity measures the degree of overlap between $R_k(i)$ and $R_k(j)$ while considering whether $x_i$ and $x_j$ are in $k$-reciprocal nearest neighbor relation. 
Since the size of the neighbor set of $x_i$ and $x_j$ would be different, we design an asymmetric Jaccard similarity as
\begin{align}
    \tilde{w}_{ij}^C =
    \begin{dcases} 
    \frac{|{R}_k(i) \cap {R}_k(j)|}{|{R}_k(i)|}, 
    & \textrm{if } j \in {R}_k(i), \\[0.3em]
    0, & \textrm{otherwise.}
    \end{dcases}
    \label{eq:Jaccard_weight}
\end{align}
%
In addition, we adopt the idea of query expansion~\cite{arandjelovic2012three,chum2007total,chum2011total} to further improve its reliability.
Specifically, $\tilde{w}_{ij}^C$ is reformulated as the average of the contextual similarities between $x_j$ and nearest neighbors of $x_i$ as follows:
\begin{equation}
\hat{w}_{ij}^C = \frac{1}{|N_{\frac{k}{2}}(i)|}\sum_{h \in N_{\frac{k}{2}}(i)} \tilde{w}_{hj}^C.
\label{eq:Jaccard_asymmetric_weight}
\end{equation}
The expanded version in Eq.~\eqref{eq:Jaccard_asymmetric_weight} is not symmetric anymore.
To ensure symmetry, the final form of the contextual similarity is defined as the average of $\hat{w}_{ij}^C$ and its transpose, which is given by
\begin{equation}
w_{ij}^{C} = w_{ji}^{C} = \frac{1}{2}(\hat{w}_{ij}^{C} + \hat{w}_{ji}^{C}).
\label{eq:Jaccard_symmetric_weight}
\end{equation}

\noindent \textbf{Contextualized semantic similarity.}
Finally, the contextualized semantic similarity between $x_i$ and $x_j$ is defined as the average of their pairwise similarity in Eq.~\eqref{eq:pairwise_similarity} and contextual similarity in Eq.~\eqref{eq:Jaccard_symmetric_weight}:
\begin{equation}
w_{ij} = \frac{1}{2}(w^{P}_{ij} + w^{C}_{ij}).
\label{eq:contextualized_semantic_similarity}
\end{equation}
Note that both of $w^{P}_{ij}$ and $w^{C}_{ij}$ lie within $[0, 1]$ thus their scales are balanced.
The contextualized semantic similarity effectively represents semantic relatedness between data by considering both of their pairwise relation and context.
We found that the contextualized semantic similarity is highly correlated with groundtruth class-equivalence, which is empirically verified in Sec.~\ref{subsec:analysis}.

\begin{table*}[!t]
\fontsize{8.5}{10.2}\selectfont
\centering
\begin{tabularx}{0.99\textwidth}{ 
  p{0.04\textwidth}>{\centering\arraybackslash}X
  >{\centering\arraybackslash}X
  >{\centering\arraybackslash}X 
  >{\centering\arraybackslash}X
  >{\centering\arraybackslash}X
  >{\centering\arraybackslash}X
  >{\centering\arraybackslash}X 
  >{\centering\arraybackslash}X
  >{\centering\arraybackslash}X
  >{\centering\arraybackslash}X
  >{\centering\arraybackslash}X}
 \toprule
\multicolumn{1}{l}{\multirow{2}{*}[-3.5mm]{Methods}}&
\multicolumn{1}{c}{\multirow{2}{*}[-3.5mm]{Arch.}}&
\multicolumn{3}{c}{CUB} & \multicolumn{3}{c}{Cars} & \multicolumn{3}{c}{SOP}\\ 
\cmidrule(lr){3-5} \cmidrule(lr){6-8} \cmidrule(lr){9-11}
& & R@1 &R@2 &R@4 &R@1 &R@2 &R@4 &R@1 &R@10 &R@100 \\ \midrule

\multicolumn{11}{l}{\emph{\textbf{Supervised}}} \\ \midrule
\multicolumn{1}{l}{ABIER~\citep{opitz2018deep}}& \multicolumn{1}{c}{G$^{512}$}  & 57.5& 69.7& 78.3& 82.0& 89.0& 93.2& 74.2 & 86.9 & 94.0 \\
\multicolumn{1}{l}{ABE~\citep{ensemble_embedding}}& \multicolumn{1}{c}{G$^{512}$}  &63.0 & 74.5& 83.3 & 85.2& 90.5& 94.0 &76.3 & 88.4 & 94.8 \\
\multicolumn{1}{l}{MS~\citep{wang2019multi}}& \multicolumn{1}{c}{BN$^{512}$}  &65.7 &77.0 & {86.3} &84.1 &90.4 &94.0 &78.2 &90.5 &96.0  \\
\multicolumn{1}{l}{PA~\citep{kim2020proxy}}& \multicolumn{1}{c}{BN$^{512}$} &{69.1} & {78.9} & 86.1 & {86.4} & {91.9} & {95.0} & {79.2} & {90.7} & {96.2}\\ 

\midrule
\multicolumn{11}{l}{\emph{\textbf{Unsupervised} (Using ImageNet pre-trained embedding networks)}} \\ \midrule
\multicolumn{1}{l}{MOM~\citep{iscen2018mining}}& \multicolumn{1}{c}{G$^{64}$}  &45.3 &57.8 & 68.6& - & - & - & - & -  & - \\
\multicolumn{1}{l}{ROUL~\citep{kan2021relative}}& \multicolumn{1}{c}{G$^{64}$}  &52.0 &64.0 & 74.5 & - & - & - & - & -  & - \\
\multicolumn{1}{l}{\ccol STML}& \multicolumn{1}{c}{\ccol G$^{64}$}  & \ccol \textbf{55.0} & \ccol \textbf{67.5} & \ccol \textbf{78.6} & \ccol \textbf{42.0} & \ccol \textbf{54.4} & \ccol \textbf{65.9} & \ccol \textbf{64.9} & \ccol \textbf{79.6} & \ccol \textbf{89.3}  \\ \midrule
\multicolumn{1}{l}{MOM~\citep{iscen2018mining}}& \multicolumn{1}{c}{G$^{128}$}  & - & - & - & 35.5 & {48.2} & {60.6} & 43.3 & 57.2 & 73.2  \\
\multicolumn{1}{l}{Examplar~\citep{dosovitskiy2015discriminative}}& \multicolumn{1}{c}{G$^{128}$}  &38.2 &50.3 & 62.8 &36.5 &48.1 &59.2 & 45.0 & 60.3 & 75.2 \\
\multicolumn{1}{l}{NCE~\citep{wu2018unsupervised}}& \multicolumn{1}{c}{G$^{128}$}  &39.2 &51.4 & 63.7 &37.5 &48.7 &59.8 & 46.6 & 62.3 & 76.8 \\
\multicolumn{1}{l}{DeepCluster~\citep{caron2018deep}}& \multicolumn{1}{c}{G$^{128}$}  &42.9 &54.1 & 65.6 &32.6 &43.8 &57.0 & 34.6 & 52.6 & 66.8  \\
\multicolumn{1}{l}{ISIF~\citep{ye2019unsupervised}}& \multicolumn{1}{c}{G$^{128}$}  &46.2 &59.0 & 70.1 &41.3 &52.3 &63.6 & 48.9 & 64.0 & 78.0 \\ 
\multicolumn{1}{l}{PSLR~\citep{ye2020probabilistic}}& \multicolumn{1}{c}{G$^{128}$}  &48.1 &60.1 & 71.8 & 43.7 & 54.8 & 66.1 & 51.1 & 66.5 & 79.8 \\ 
\multicolumn{1}{l}{ROUL~\citep{kan2021relative}}& \multicolumn{1}{c}{G$^{128}$}  & 56.7 & 68.4 & 78.3 & 45.0 & 56.9 & 68.4 & 53.4 & 68.8 & 81.7 \\ 
\multicolumn{1}{l}{SAN~\citep{li2021spatial}}& \multicolumn{1}{c}{G$^{128}$}  & 55.9 & 68.0 & 78.6 & 44.2 & 55.5 & 66.8 & 58.7 & 73.1 & 84.6 \\ 
\multicolumn{1}{l}{\ccol STML} & \multicolumn{1}{c}{\ccol G$^{128}$}   &\ccol \textbf{59.7} &\ccol \textbf{71.2} &\ccol \textbf{81.0} &\ccol \textbf{49.0} &\ccol\textbf{60.4} &\ccol \textbf{71.3} &\ccol \textbf{65.8} &\ccol \textbf{80.1} &\ccol\textbf{89.9} \\ \midrule
\multicolumn{1}{l}{UDML-SS~\citep{cao2019unsupervised}}& \multicolumn{1}{c}{G$^{512}$}  &54.7 & 66.9 & 77.4 &45.1 & 56.1 & 66.5 & 63.5 & 78.0 & 88.6 \\
\multicolumn{1}{l}{TAC-CCL~\citep{li2020unsupervised}}& \multicolumn{1}{c}{G$^{512}$}  &57.5 & 68.8 & 78.8 &46.1 & 56.9 & 67.5 & 63.9 & 77.6 & 87.8  \\
\multicolumn{1}{l}{UHML~\citep{yan2021unsupervised}}& \multicolumn{1}{c}{G$^{512}$}  &58.9 & 70.6 & 80.4 &47.7 & 58.9 & 70.3 & 65.1 & 78.2 & 88.3  \\
\multicolumn{1}{l}{\ccol STML} & \multicolumn{1}{c}{\ccol G$^{512}$}   &\ccol \textbf{60.6} &\ccol \textbf{71.7} &\ccol \textbf{81.5} &\ccol \textbf{50.5} &\ccol \textbf{61.8} &\ccol \textbf{71.7} &\ccol \textbf{65.3} &\ccol \textbf{79.8} &\ccol \textbf{89.8} \\ 
\midrule
\multicolumn{1}{l}{UDML-SS~\cite{cao2019unsupervised}} & \multicolumn{1}{c}{{BN}$^{512}$}   & 63.7 & 75.0 & 83.8 & - & - & - & - & - & - \\ 
\multicolumn{1}{l}{\ccol STML} & \multicolumn{1}{c}{\ccol {BN}$^{512}$}   &\ccol \textbf{68.0} &\ccol \textbf{78.8} &\ccol \textbf{86.4} &\ccol \textbf{66.2} &\ccol \textbf{74.5} &\ccol \textbf{81.9} &\ccol \textbf{69.7} &\ccol \textbf{82.7} &\ccol \textbf{91.2} \\ 
\bottomrule
\end{tabularx}
\vspace{-1mm}
\caption{
Performance of the unsupervised and supervised metric learning methods on the three datasets. 
Their network architectures are denoted by abbreviations, G--GoogleNet~\cite{Szegedy2015}, BN--Inception with BatchNorm~\cite{Batchnorm}, where superscripts denote their embedding dimensions. 
}
\label{tab:dml_unsup}
\vspace{-1mm}
\end{table*}

\section{Experiments}
\label{sec:experiments}

This section validates the effectiveness of STML on two metric learning tasks with lack of supervision, \ie, unsupervised metric learning and semi-supervised metric learning.

\subsection{Unsupervised Metric Learning}
We consider the student embedding model trained by STML as our final model, 
which is evaluated and compared with state-of-the-art unsupervised metric learning methods and representative self-supervised learning methods.

\subsubsection{Experimental Setup}
\label{subsec:experiment_unsup_detail}
\noindent \textbf{Datasets and evaluation.}
The models are evaluated and compared on three benchmark datasets, \ie, CUB-200-2011 (CUB)~\cite{CUB200}, Cars-196 (Cars)~\cite{krause20133d}, and Stanford Online Product (SOP)~\cite{songCVPR16}; 
for their train-test splits, we directly follow the standard protocol presented in~\cite{ye2019unsupervised}.
Performance on these datasets is evaluated by Recall@$k$, the fraction of queries that have at least one relevant sample in their $k$-nearest neighbors on a learned embedding space.

\noindent \textbf{Embedding networks.}
For comparisons to previous work, we employ GoogleNet~\cite{Szegedy2015} and Inception-BN~\cite{Batchnorm} as our teacher and student embedding model architecture.
The dimension of embedding layers $g_t$ and $g_s$ are set as the output size of the last pooling, and that of $f_s$ follows each experimental setting.
We append an $l_2$ normalization layer on top of the teacher so that the Euclidean distances in Eq.~\eqref{eq:pairwise_similarity} are bounded for computing semantic similarities stably.

\noindent \textbf{Implementation details.}
The student model of STML is optimized by AdamP with Nesterov momentum~\cite{heo2021adamp} for 90 epochs on a single Titan RTX; the learning rate is initialized to $10^{-4}$ and scaled down by the cosine decay function~\cite{loshchilov2016sgdr}.
Training images are randomly cropped to $227 \times 227$ and flipped horizontally at random while test images are resized to $256 \times 256$ then center cropped. 
For all experiments, $\sigma$ in Eq.~\eqref{eq:smooth_contrastive_mu} is set to $3$.
The margin $\delta$, momentum coefficient $m$ and the number of nearest neighbors $k$ are $1.0$, $0.999$ and $10$ on the CUB and Cars datasets, and $0.9$, $0.9$ and $4$ on the SOP dataset.
Note that $k$ in the nearest neighbor batch construction and that in Eq.~\eqref{eq:Reciprocal_knn} are identical.

\noindent \textbf{Batch construction.}
Nearest neighbors used to construct mini-batches in STML are computed on the student embedding space and update at every epoch.
On the CUB and Cars datasets, each mini-batch consists of $120$ images including $24$ queries and their $4$ nearest neighbors. 
Since the SOP dataset has a tiny number of samples per class, we construct a mini-batch with $60$ queries and their top-$1$ neighbors on the dataset.
Also, all images are randomly augmented twice following the multi-view augmentation strategy in~\cite{kim2021embedding}.
\vspace{-2mm}

\subsubsection{Quantitative Results}

For fair comparisons to previous work, performance evaluation is done in the following settings:
GoogleNet with 64/128/512 embedding dimensions, and Inception with BatchNorm with 512 embedding dimensions.
All backbone networks in this experiment are pre-trained on ImageNet. 
The results on the three datasets are summarized in Table~\ref{tab:dml_unsup}. 

\begin{table}
\centering
\fontsize{8.5}{10.5}\selectfont
\begin{tabularx}{0.48 \textwidth}{
   p{0.11\textwidth}>{\centering\arraybackslash}X
   >{\centering\arraybackslash}X
   >{\centering\arraybackslash}X
   >{\centering\arraybackslash}X}
\toprule
\multicolumn{1}{l}{\multirow{2}{*}[-3.5mm]{Methods}}&
\multicolumn{3}{c}{SOP} \\  \cmidrule(lr){2-4}
&R@1 &R@10 &R@100\\ \midrule
\multicolumn{1}{l}{NCE~\citep{wu2018unsupervised}}  &34.4 &49.0 & 65.2   \\
\multicolumn{1}{l}{ISIF~\citep{ye2019unsupervised}}  &39.7 &54.9 & 71.0  \\
\multicolumn{1}{l}{PSLR~\citep{ye2020probabilistic}}  &42.3 &57.7 & 72.5  \\
\multicolumn{1}{l}{ROUL~\citep{kan2021relative}}  &45.4 &60.5 & 74.8  \\
\multicolumn{1}{l}{SAN~\citep{li2021spatial}}  &46.3 & 61.9 & 77.0  \\
\multicolumn{1}{l}{\ccol STML} & \ccol \textbf{60.7} & \ccol\textbf{74.8}  & \ccol\textbf{85.2}  \\ \bottomrule
\end{tabularx}
\vspace{-1mm}
\caption{
Performance on the SOP dataset using ResNet18 without pre-trained weights.
}
\label{tab:dml_scratch}
\end{table}

On all the datasets, our model achieves state-of-the-art performance in every setting. It improves the previous best score by a large margin, from 3.8\% to 7.1\% at Recall@1 with the 128 embedding dimensions. We note that our model clearly surpasses all existing models whose embedding dimension is 4 times higher on all the datasets, sometimes at 8 times higher. Moreover, on the CUB dataset, it even outperforms some of the supervised learning methods, \eg, ABIER~\cite{opitz2018deep} and MS~\cite{wang2019multi}, given the same embedding networks. These results suggest that our framework allows the model to learn compact yet effective embedding spaces by providing high-quality synthetic supervision.
Last but not least, STML enables more efficient learning than existing pseudo labeling methods due to no use of off-the-shelf algorithms, especially on large-scale SOP dataset.

\subsubsection{Unsupervised Metric Learning from Scratch}
We demonstrate that STML is also effective for 
metric learning from scratch without pre-training on ImageNet.
Following~\citep{ye2019unsupervised}, we employ randomly initialized ResNet18~\cite{resnet} with 128 embedding dimensions as the backbone network, and evaluate our model trained on the SOP dataset.
As shown in Table~\ref{tab:dml_scratch}, our model substantially outperforms existing methods including those for self-supervised representation learning in this setting. 
This result suggests that STML can generate reliable pseudo labels even on a randomly initialized embedding space, and our model trained by STML well generalizes to unseen class.

\begin{figure} [!t]
\centering
\includegraphics[width = 1.0\columnwidth]{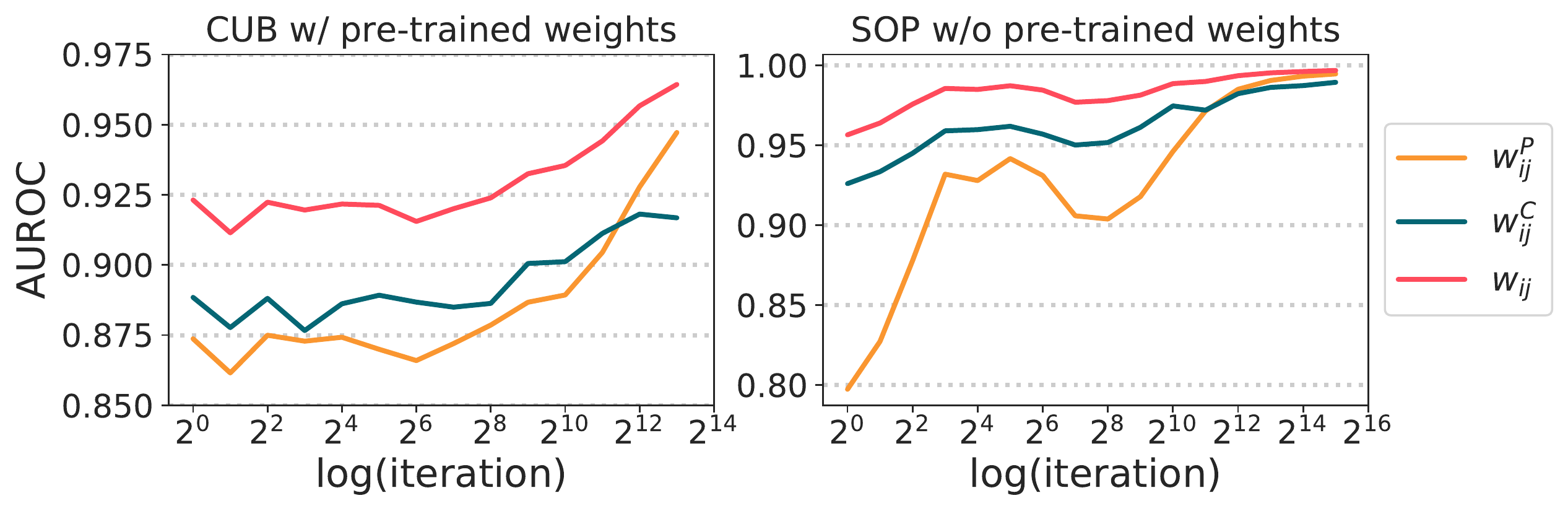}
\vspace{-6mm}
\caption{
Comparison between the contextualized semantic similarity and its two components in terms of class-equivalence prediction accuracy on the CUB dataset with pre-trained weights~(\emph{left}) and SOP dataset without pre-trained weights~(\emph{right}).
} 
\label{fig:ROC_curve}
\end{figure}
%
%
\begin{table}
\centering
\fontsize{8.5}{10.5}\selectfont
\begin{tabularx}{0.48 \textwidth}{ 
   p{0.14\textwidth}>{\centering\arraybackslash}X
   >{\centering\arraybackslash}X
   >{\centering\arraybackslash}X}
 \toprule
\multicolumn{1}{l}{\multirow{2}{*}[-3.5mm]{Methods}}&
\multicolumn{1}{c}{CUB} & \multicolumn{1}{c}{Cars} \\ \cmidrule(lr){2-2} \cmidrule(lr){3-3}
&R@1 &R@1 \\ \midrule
\multicolumn{1}{l}{\ccol STML}& \ccol \textbf{60.6} & \ccol \textbf{50.5} \\
\multicolumn{1}{l}{w/o contextual pairwise similarity}&  {44.2} &  {40.2} \\
\multicolumn{1}{l}{w/o pairwise similarity}&  {56.0} & {45.0} \\
\multicolumn{1}{l}{w/o relaxed contrastive loss}&  {53.4} &  {38.7} \\
\multicolumn{1}{l}{w/o momentum update}&  {56.8} &  {46.1} \\
\multicolumn{1}{l}{w/o NN batch construction}&  {54.6} & {44.8} \\
\multicolumn{1}{l}{w/o KL divergence}&  {59.2} & {49.4} \\
\bottomrule
\end{tabularx}
\vspace{-1mm}
\caption{
Ablation study of the components in STML on the CUB and Cars datasets.
}
\label{tab:dml_ablation}
\end{table}

\subsubsection{In-depth Analysis}
\label{subsec:analysis}
\noindent \textbf{Correlation between similarities and class-equivalence.}
To justify the contextualized semantic similarity, we compare our three similarity metrics (\ie, pairwise similarity $w^P_{ij}$, contextual similarity $w^C_{ij}$, contextualized semantic similarity $w_{ij}$) in terms of how well they are correlated with class-equivalence, \ie, the groundtruth for supervised metric learning.
Specifically, we predict class-equivalence of every pair of data by thresholding its semantic similarities, and count the accuracy of such predictions at various threshold settings.
Fig.~\ref{fig:ROC_curve} presents AUROC scores of our contextualized semantic similarity and its two components during training. 
The results suggest that the accuracy of the pairwise similarity relies heavily on the quality of the teacher embedding space,
while the contextual similarity helps infer class-equivalence correctly even at early stages of training without pre-trained weights.
The contextualized semantic similarity achieves the best consistently, which also indicates that the pairwise and contextual similarities are complementary to each other.

\noindent \textbf{Ablation study.}
The impact of each component in STML is reported in Table~\ref{tab:dml_ablation}.
The results in the table are obtained by using GoogleNet with 512 embedding dimensions.
The analysis reveals that the contextual similarity contributes most to the performance while the pairwise similarity has the minimum impact; Fig.~\ref{fig:ROC_curve} also supports this observation.
Even if both similarities are used, the performance is largely degraded when using the conventional contrastive loss~\cite{Hadsell2006} instead of the relaxed version since
the conventional one quantizes our semantic similarity into a binary value, leading to significant information loss.
The momentum-based update of the teacher model and the nearest neighbor batch construction strategy also have significant impacts;
the performance degrades substantially when the teacher model is updated directly by the student counterpart or the batch construction strategy is replaced with a random sampling. The KL divergence loss has relatively little influence compared to other components.

\begin{figure} [!t]
\centering
\includegraphics[width = 0.95\columnwidth]{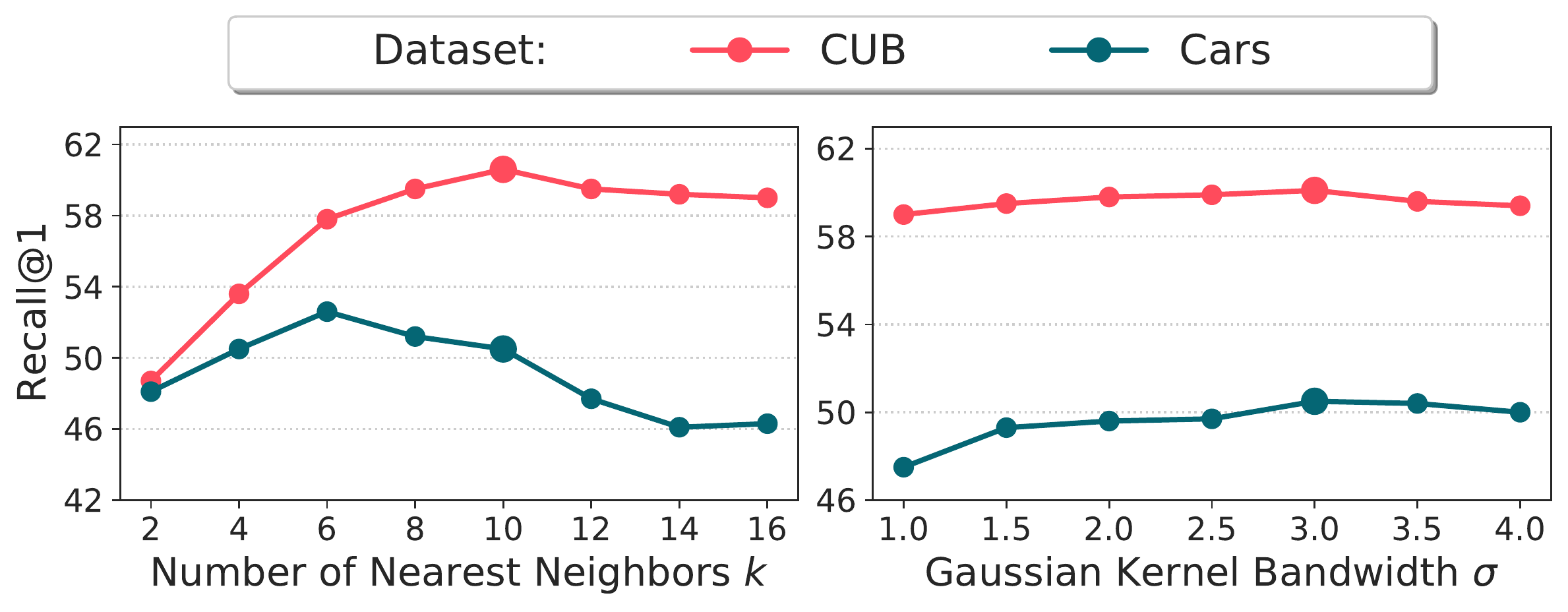}
\vspace{-2mm}
\caption{Recall@1 versus the hyper-parameters $k$ and $\sigma$ on the CUB and Cars datasets. Larger markers indicate our default experimental setting.
}
\label{fig:hypparam_sensitivity}
\end{figure}

\begin{table}
\centering
\fontsize{8.5}{10.5}\selectfont
\begin{tabularx}{0.48 \textwidth}{ 
  p{0.07\textwidth}>{\centering\arraybackslash}X
  >{\centering\arraybackslash}X
  >{\centering\arraybackslash}X
  >{\centering\arraybackslash}X}
 \toprule
\multicolumn{1}{l}{\multirow{2}{*}[-3.5mm]{Methods}}&
\multicolumn{1}{c}{\multirow{2}{*}[-3.5mm]{Arch.}}&
\multicolumn{1}{c}{CUB} & \multicolumn{1}{c}{Cars} & \multicolumn{1}{c}{SOP} \\ \cmidrule(lr){3-3} \cmidrule(lr){4-4} \cmidrule(lr){5-5}
& &R@1 &R@1 &R@1 \\ \midrule
\multicolumn{1}{l}{MoCo~\cite{he2020momentum}}& \multicolumn{1}{c}{G$^{128}$} &{48.3} & {37.2} & {53.3} \\
\multicolumn{1}{l}{BYOL~\cite{grill2020bootstrap}}& \multicolumn{1}{c}{G$^{128}$} &{47.7} & {31.5} & {48.7} \\
\multicolumn{1}{l}{MSF~\cite{koohpayegani2021mean}}&  {G}$^{128}$ &  {46.0} &  {30.5} &  {41.6} \\
\multicolumn{1}{l}{\ccol STML}&  \multicolumn{1}{c}{\ccol G$^{128}$} &\ccol \textbf{59.7} & \ccol \textbf{49.0} & \ccol \textbf{65.8} \\ \hline
\multicolumn{1}{l}{MoCo~\cite{he2020momentum}}& \multicolumn{1}{c}{G$^{512}$} &{51.0} & {39.0} & {53.4} \\
\multicolumn{1}{l}{BYOL~\cite{grill2020bootstrap}}& \multicolumn{1}{c}{G$^{512}$} &{50.9} & {38.5} & {51.2} \\
\multicolumn{1}{l}{MSF~\cite{koohpayegani2021mean}}&  {G}$^{512}$ &  {49.6} &  {32.5} &  {47.8} \\
\multicolumn{1}{l}{\ccol STML}&  \multicolumn{1}{c}{\ccol G$^{512}$} &\ccol \textbf{60.6} & \ccol \textbf{50.5} & \ccol \textbf{65.3} \\
\bottomrule
\end{tabularx}
\vspace{-1mm}
\caption{
Comparison between STML and recent self-supervised representation learning methods using momentum encoder.
}
\label{tab:self_sup}
\end{table}

\begin{table*}[!t]
\fontsize{8.2}{9.5}\selectfont
\centering
\begin{tabularx}{1\textwidth}{ 
   p{0.1\textwidth}>{\centering\arraybackslash}X
   >{\centering\arraybackslash}X
   >{\centering\arraybackslash}X
   >{\centering\arraybackslash}X
   >{\centering\arraybackslash}X
   >{\centering\arraybackslash}X
   >{\centering\arraybackslash}X
   >{\centering\arraybackslash}X
   >{\centering\arraybackslash}X}
 \toprule
\multicolumn{1}{l}{\multirow{2}{*}[-3.5mm]{Methods}}&
\multicolumn{1}{c}{\multirow{2}{*}[-3.5mm]{Arch.}}&
\multicolumn{1}{c}{\multirow{2}{*}[-3.5mm]{Init.}}&
\multicolumn{3}{c}{CUB} & \multicolumn{3}{c}{Cars} \\ \cmidrule(lr){4-6} \cmidrule(lr){7-9}
& & &MAP@R &RP &R@1 &MAP@R &RP &R@1 \\ \midrule

\multicolumn{9}{l}{\emph{\textbf{Supervised}}} \\ \midrule
\multicolumn{1}{l}{Contrastive~\citep{Chopra2005}}& \multicolumn{1}{c}{R50$^{512}$}& \multicolumn{1}{c}{ImageNet} &25.0 &35.8 & 65.3 & 26.0 &36.4 &81.2 \\
\multicolumn{1}{l}{MS~\citep{wang2019multi}}& \multicolumn{1}{c}{R50$^{512}$}& \multicolumn{1}{c}{ImageNet} &26.4 &37.5 & 66.3 & 28.3 &38.3 &85.2  \\
\multicolumn{1}{l}{Proxy-Anchor~\citep{kim2020proxy}}& \multicolumn{1}{c}{R50$^{512}$}& \multicolumn{1}{c}{ImageNet} &28.3 &39.1 & 69.9 & 30.5 & 39.9 &87.7 \\ 
\multicolumn{1}{l}{Contrastive~\citep{Chopra2005}}& \multicolumn{1}{c}{R50$^{512}$}& \multicolumn{1}{c}{SwAV~\cite{caron2020unsupervised}} &29.3 & 39.8 & 71.2 & 31.7 &41.2& 88.1\\
\multicolumn{1}{l}{MS~\citep{wang2019multi}}& \multicolumn{1}{c}{R50$^{512}$}& \multicolumn{1}{c}{SwAV~\cite{caron2020unsupervised}} &29.2 & 40.2 & 70.8 & 33.4 & 42.7 & 89.3  \\
\multicolumn{1}{l}{Proxy-Anchor~\citep{kim2020proxy}}& \multicolumn{1}{c}{R50$^{512}$}& \multicolumn{1}{c}{SwAV~\cite{caron2020unsupervised}} &31.7 &42.0 & 74.6 & 35.1 &44.1 &90.4 \\ 

\midrule
\multicolumn{9}{l}{\emph{{\textbf{Semi-supervised} (Labeled dataset +  Additional unlabeled dataset)}}} \\ \midrule
\multicolumn{1}{l}{SLADE (MS)~\citep{duan2020slade}}& \multicolumn{1}{c}{R50$^{512}$}& \multicolumn{1}{c}{ImageNet} &30.9 & 41.9 & 69.6 &32.1 & 41.5 & 87.4 \\ 
\multicolumn{1}{l}{\ccol STML (MS)} & \multicolumn{1}{c}{\ccol {R50}$^{512}$} & \multicolumn{1}{c}{\ccol ImageNet} &\ccol \textbf{37.8} &\ccol \textbf{47.9} &\ccol \textbf{75.5} &\ccol \textbf{38.4} &\ccol \textbf{46.1} &\ccol \textbf{93.0}  \\ \midrule

\multicolumn{1}{l}{SLADE (MS)~\citep{duan2020slade}}& \multicolumn{1}{c}{R50$^{512}$}& \multicolumn{1}{c}{SwAV~\cite{caron2020unsupervised}} &33.9 &44.4 & 74.1 &38.0 &\textbf{46.9} &91.5 \\ 
\multicolumn{1}{l}{\ccol STML (MS)} & \multicolumn{1}{c}{\ccol {R50}$^{512}$} & \multicolumn{1}{c}{\ccol SwAV~\cite{caron2020unsupervised}} &\ccol \textbf{35.2} &\ccol \textbf{45.3} &\ccol \textbf{76.2} &\ccol \textbf{38.2} &\ccol{46.5} &\ccol \textbf{93.2}  \\ 

\bottomrule
\end{tabularx}
\vspace{-1mm}
\caption{
Performance of the supervised and semi-supervised methods on the two datasets. 
Network architectures of the methods are all  ResNet50 (R50)~\cite{resnet} and superscripts denote their embedding dimensions.
Also, the column ``Init.'' indicates whether the models are pre-trained on ImageNet or by SwAV.
Our model and SLADE are all fine-tuned with the MS loss.
}
\label{tab:dml_semi}
\vspace{-1mm}
\end{table*}

\noindent \textbf{Effect of hyper-parameters.}
We investigate how much sensitive STML is to the hyper-parameters $k$ and $\sigma$.
Specifically, we measure performance of STML employing GoogleNet with 512 embedding dimensions on the CUB and Cars datasets while varying values of $k$ and $\sigma$.
The results shown in Fig.~\ref{fig:hypparam_sensitivity} suggest that the performance of STML is consistently high when $k<14$, and $\sigma$ has virtually no effect on the performance if $\sigma>1$ on both datasets.
Moreover, whenever $k<14$ our model outperforms all existing unsupervised metric learning methods. Note that our final results reported in Table~\ref{tab:dml_unsup} are not the best scores in Fig.~\ref{fig:hypparam_sensitivity} since we did not tune the hyper-parameters on the test splits.

\noindent \textbf{Comparison with recent SSL methods.}
STML has components used in recent self-supervised representation learning (SSL) methods such as the momentum encoder. To validate that the improvements of our framework do not originate from its structure, such methods are compared with STML under the same setting. Table~\ref{tab:self_sup} shows that MoCo and BYOL clearly underperform our method in all settings and achieve similar performance to unsupervised metric learning methods based on instance-level surrogate classes. Although they are structurally similar to STML, the reason for the degradation of performance is that they cannot capture intra-class variation unlike our method, and are limited in representing semantic similarities between data.

\subsection{Semi-supervised Metric Learning}
Semi-supervised metric learning aims to improve performance of supervised metric learning by exploiting additional unlabeled data~\citep{duan2020slade}.
STML can be employed for this task naturally since it provides a way of using the unlabeled data for training by predicting their class-equivalence relations as synthetic supervision. 
In this section, we validate the efficacy of STML for semi-supervised metric learning and compare it with SLADE~\citep{duan2020slade}, the first and only existing method for the task, on the CUB and Cars datasets.

\subsubsection{Experimental Setup}
\vspace{-1mm}
\label{subsec:experiment_semisup_detail}
\noindent \textbf{Datasets and evaluation.}
We directly adopt the evaluation protocol of SLADE.
Specifically, we employ two combinations of labeled and unlabeled datasets, \{CUB~\citep{CUB200}, NABirds~\citep{van2015building}\} and \{Cars~\citep{krause20133d}, CompCars~\citep{yang2015large}\}, and three performance metrics, MAP@R, RP, and Recall@1.

\noindent \textbf{Embedding networks.}
For comparisons to SLADE, we employ ResNet50~\citep{resnet} with 512 embedding dimensions as the backbone network of the teacher and student models. 
Before applying STML, these models are pre-trained either on ImageNet~\citep{Imagenet} or by SwAV~\citep{caron2020unsupervised}, then fine-tuned with the MS loss~\citep{wang2019multi} on a labeled dataset (\ie, CUB or Cars).

\noindent \textbf{Implementation details.}
In this task, STML adopts an embedding model trained on a labeled dataset as the initial teacher and student models and trains them using both labeled and unlabeled data.
Note that it computes and utilizes contextualized semantic similarities as supervision for labeled data as well as unlabeled data.
For all experiments, $m$ in Eq.~\eqref{eq:teacher_momentum_update} and $\sigma$ in Eq.~\eqref{eq:contextualized_semantic_similarity} are set to 0.9999 and 2, respectively, and the rest settings are the same as those for unsupervised metric learning.

\subsubsection{Results}
STML is compared with SLADE and state-of-the-art supervised metric learning methods on the two datasets.
The results summarized in Table~\ref{tab:dml_semi} show that STML improves performance of its supervised learning counterpart (\ie, MS) significantly regardless of the model initialization schemes.
In particular, it enhances the performance by more than 5\% in Recall@1 on both datasets for ImageNet pre-trained setting.
Also, STML clearly outperforms SLADE in all the settings except one, where their gap is only 0.4\% in RP. 

\section{Conclusion}
\label{sec:conclusion}

We have proposed STML, a novel end-to-end unsupervised metric learning framework that estimates and exploits semantic relations between samples as pseudo labels.
The pseudo labels in STML are estimated by investigating contexts of data as well as their pairwise distances on an embedding space.
STML has achieved the state of the art on the three benchmark datasets for unsupervised metric learning, and demonstrated impressive performance even without pre-trained weights. 
Moreover, it has been applied to semi-supervised learning, and outperformed state of the art on the two benchmark datasets.
Unfortunately, like other approaches, STML has a limitation in that it uses ImageNet pre-trained weights in some settings. In the future, we will explore extensions of our method where class labels are never involved in the training.

\vspace{1mm}
{\small
\noindent \textbf{Acknowledgement.} 
This work was supported by 
\mbox{the NRF~grant}, 
the IITP grant, 
and R\&D program for Advanced Integrated-intelligence for IDentification, 
\mbox{funded by Ministry of Science} and ICT, Korea
(NRF-2021R1A2C3012728, 
 NRF-2018R1A5-A1060031, 
 NRF-2018M3E3A1057306, 
 No.2019-0-01906 Artificial Intelligence Graduate School Program--POSTECH). 
}

{\small
\bibliographystyle{ieee_fullname}
\bibliography{cvlab_kwak}
}

\renewcommand\thesection{\Alph{section}}
\setcounter{section}{0}

\newpage
\bigskip~\bigskip~\bigskip~\bigskip~\bigskip~\bigskip~\bigskip~\bigskip~\bigskip~\bigskip~\bigskip~\bigskip~\bigskip~\bigskip~\bigskip~\bigskip~\bigskip~\bigskip~\bigskip~\bigskip~\bigskip~\bigskip~\bigskip~\bigskip~\bigskip~\bigskip~\bigskip~\bigskip~\bigskip~\bigskip~\bigskip~\bigskip~\bigskip~\bigskip~\bigskip~\bigskip~\bigskip~\bigskip~\bigskip~\bigskip~\bigskip~\bigskip~\bigskip~\bigskip~\bigskip~\bigskip~\bigskip~\bigskip~\bigskip~\bigskip~\bigskip~\bigskip~\bigskip~\bigskip~\bigskip~\bigskip~\bigskip~\bigskip~\bigskip~\bigskip~\bigskip~\bigskip~\bigskip

\section{Appendix}
This appendix provides further analyses, and additional experimental results, all of which are left out from the main paper due to the space limit. 
In Section~\ref{sec:further_analysis}, we first discuss properties of the contextualized semantic similarity in detail.
Section~\ref{sec:qualitative_results} presents $t$-SNE visualizations of the learned embedding space and qualitative examples of image retrieval on the three benchmark datasets.

\subsection{In-depth Analysis on Contextualized Semantic Similarity}
\label{sec:further_analysis}

\begin{figure*} [!t]
\centering
\includegraphics[width = 0.61\textheight]{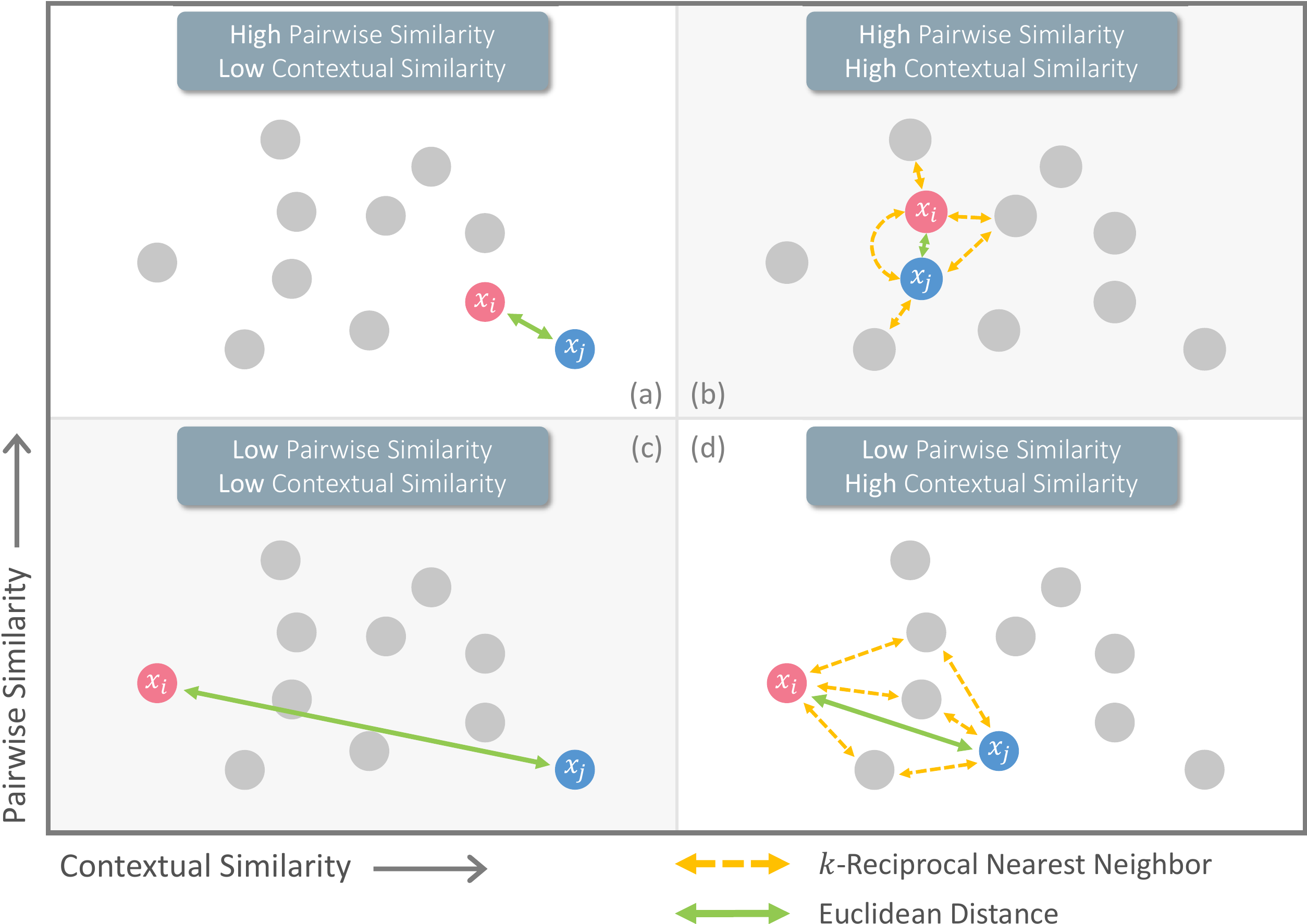}
\caption{
Difference in contextualized semantic similarity according to the relation between two samples.
} 
\label{fig:similarity_comparisions}
\end{figure*}

STML employs the contextualized semantic similarity as pseudo label that can capture semantic relations between data.
%
To better understand its property, we illustrate in Fig.~\ref{fig:similarity_comparisions} how the contextualized semantic similarity is determined in four different cases of relations between samples.
As shown in Fig.~\ref{fig:similarity_comparisions}(b), when a pair of samples not only share many $k$-reciprocal nearest neighbors but also have a small Euclidean distance, they obviously have high contextualized semantic similarity; 
they are visually similar and placed on the same manifold, thus highly likely to belong to the same class.
In contrast, as shown in Fig.~\ref{fig:similarity_comparisions}(c), two samples that are far apart and share no neighbor have a low contextualized semantic similarity as they are semantically unrelated; we found that relations of most pairs of unlabeled data fall into this case.
In Fig.~\ref{fig:similarity_comparisions}(a) and \ref{fig:similarity_comparisions}(d), where only one of the pairwise and contextual similarities is high, two samples have a higher contextualized semantic similarity than that in the case of Fig.~\ref{fig:similarity_comparisions}(c), but lower than that in the case of Fig.~\ref{fig:similarity_comparisions}(b). 
%
%
Note that although the class-equivalence between samples is uncertain in the cases of Fig.~\ref{fig:similarity_comparisions}(a) and \ref{fig:similarity_comparisions}(d), the pairwise and contextual similarities could be overly high in each of these cases, leading to noisy synthetic supervision.
On the other hand, the contextualized semantic similarity is modest thus provides reliable pseudo labels in these cases.

\begin{figure*} [!t]
\centering
\includegraphics[width = 0.61\textheight]{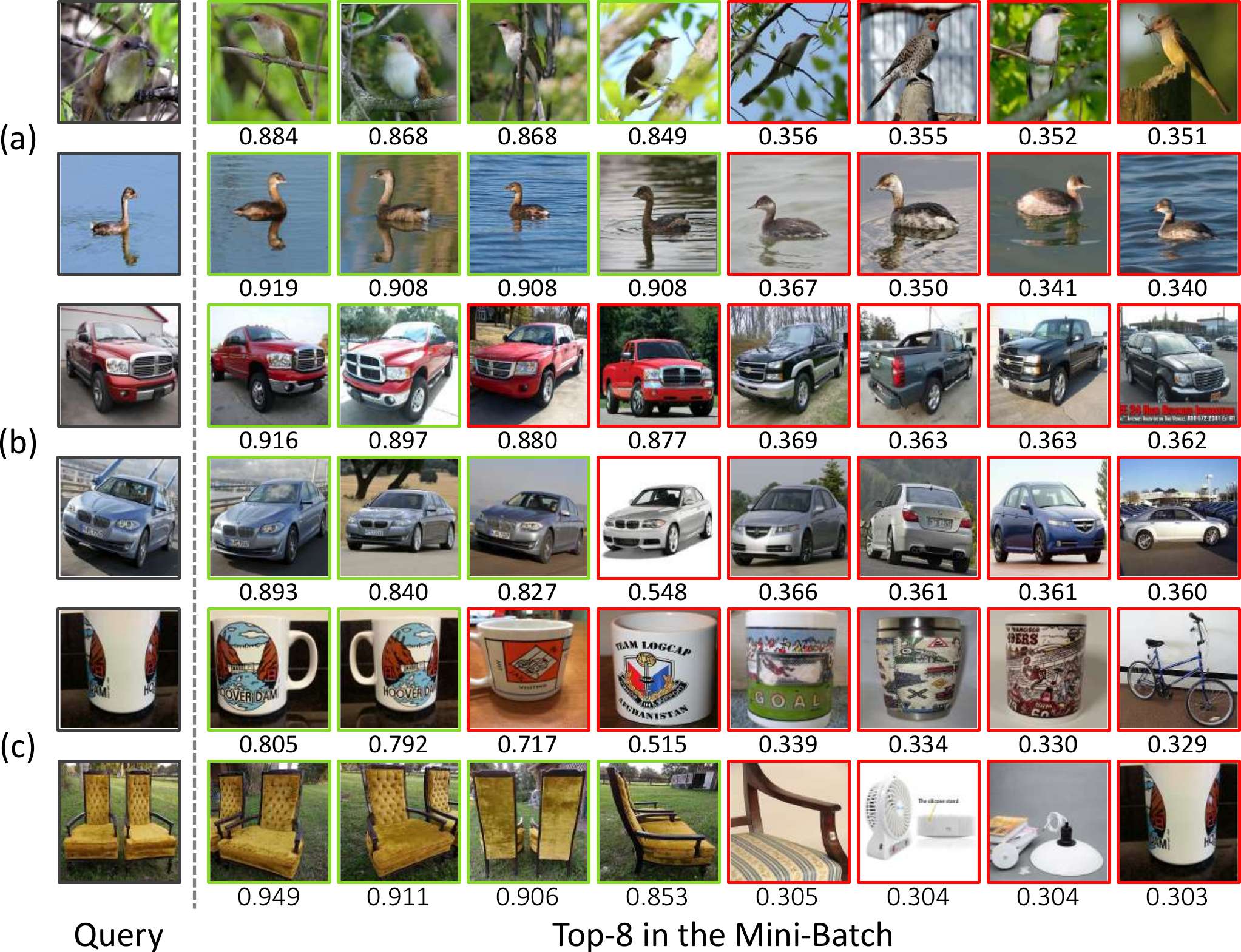}
\caption{
Image pairs sorted by their contextualized semantic similarity on (a) CUB-200-2011, (b) Cars-196, and (c) SOP datasets. Images with green boundary are of the same class as query and those with red boundary are of a different class from query.
} 
\label{fig:css_sort}
\end{figure*}

The contextualized semantic similarity allows the relaxed contrastive loss to exploit reliable supervision.
To demonstrate this, we empirically analyze the information provided by the contextualized semantic similarity.
Fig.~\ref{fig:css_sort} presents the top-8 samples in a mini-batch of each query in terms of their contextualized semantic similarity on the three benchmark datasets~\cite{krause20133d,songCVPR16,CUB200}. 
The results are obtained by the source embedding model where the number of nearest neighbor $k$ is $5$. 
As shown in the figure, the contextualized semantic similarity is highly correlated with the groundtruth class-equivalence on all datasets. 
In particular, birds of the same class are assigned higher contextualized semantic similarities although all birds are floating on water with similar poses in the 2nd row of Fig.~\ref{fig:css_sort}.
Moreover, as can be seen in the 1st and 6th rows of Fig.~\ref{fig:css_sort},
the contextualized semantic similarity successfully captures class-equivalence even under view-point variations.

\subsection{Qualitative Results}
\label{sec:qualitative_results}

\begin{figure*} [!t]
\centering
\includegraphics[width = 0.89 \textwidth]{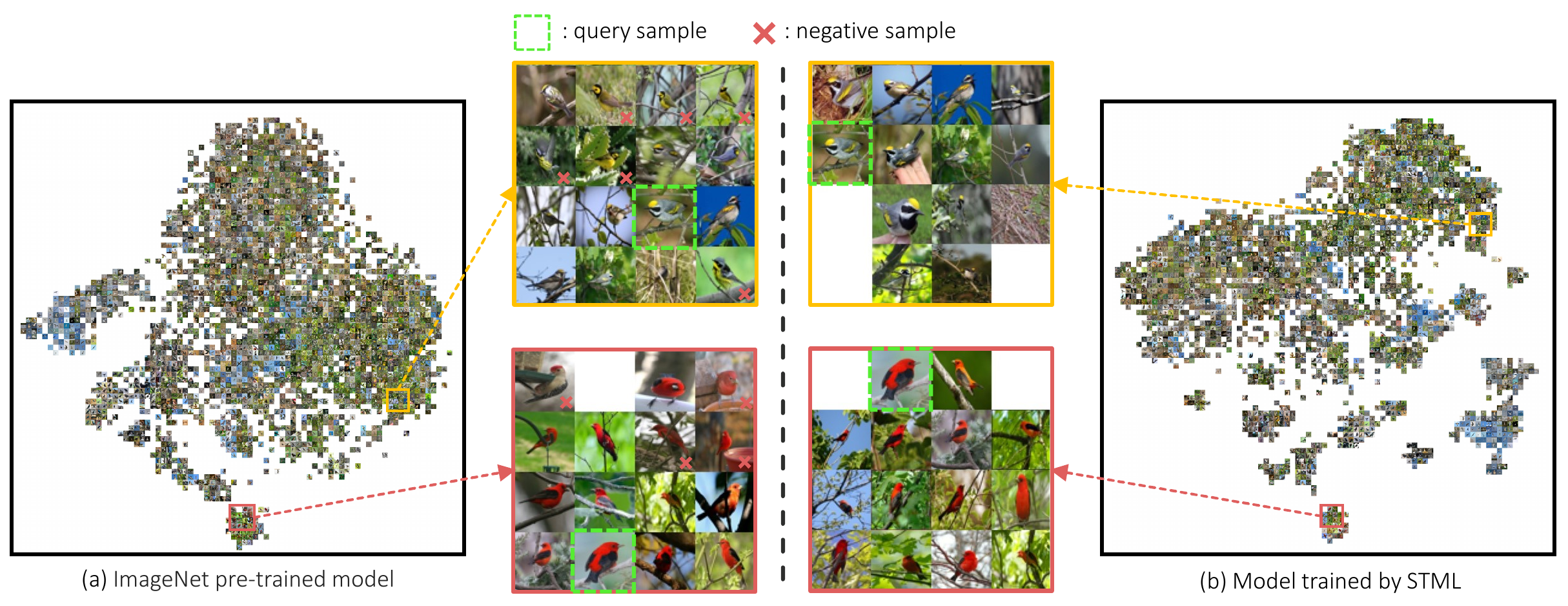}
\caption{
$t$-SNE visualization of (a) ImageNet pre-trained model and (b) model trained by STML on the test split of CUB-200-2011 dataset. A green dotted box indicates query samples, and a red cross mark indicates samples with different class labels from the query.
} 
\label{fig:tsne_vis_cub}
\end{figure*}

\begin{figure*} [!t]
\centering
\includegraphics[width = 0.89 \textwidth]{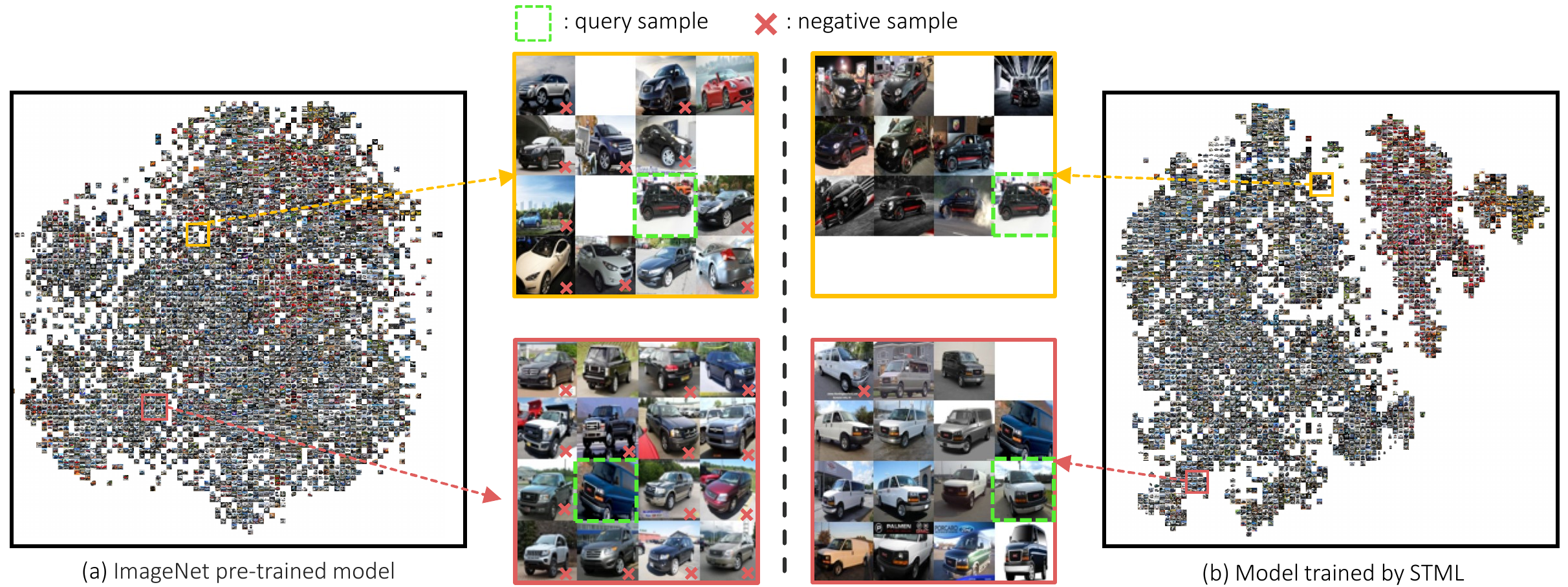}
\caption{
$t$-SNE visualization of (a) ImageNet pre-trained model and (b) model trained by STML on the test split of Cars-196 dataset. A green dotted box indicates query samples, and a red cross mark indicates samples with different class labels from the query.
} 
\label{fig:tsne_vis_cars}
\end{figure*}

\begin{figure*} [!t]
\centering
\includegraphics[width = 0.89 \textwidth]{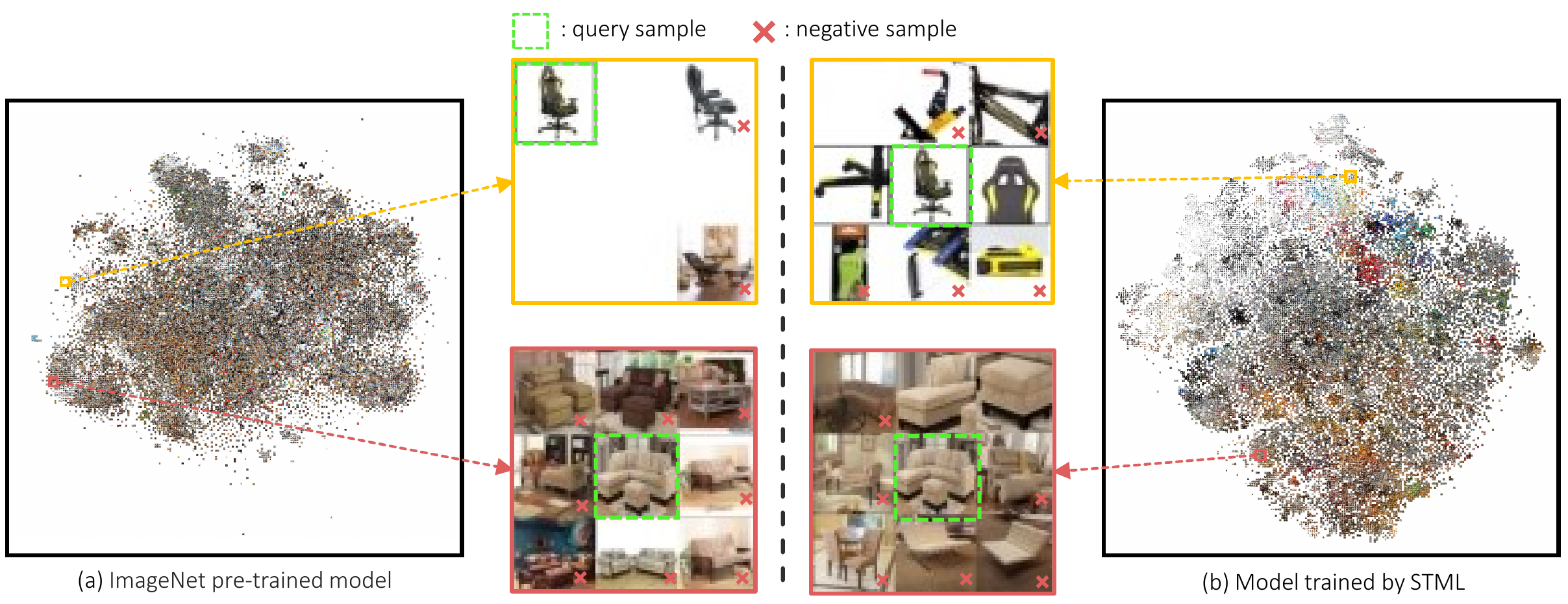}
\caption{
$t$-SNE visualization of (a) ImageNet pre-trained model and (b) model trained by STML on the test split of SOP dataset. A green dotted box indicates query samples, and a red cross mark indicates samples with different class labels from the query.
} 
\label{fig:tsne_vis_sop}
\end{figure*}

To demonstrate the efficacy of the proposed method empirically, we present Barnes-Hut $t$-SNE visualizations~\cite{vandermaaten14a} on the CUB-200-2011~\cite{CUB200}, Cars-196~\cite{krause20133d} and SOP~\cite{songCVPR16} datasets in Fig.~\ref{fig:tsne_vis_cub}, \ref{fig:tsne_vis_cars}, and \ref{fig:tsne_vis_sop}, respectively.
The visualizations of the embedding space learned by STML are compared with those of the embedding space before training (\ie, ImageNet pre-trained) using GoogleNet~\cite{Szegedy2015} with 128 embedding dimensions. As shown in Fig.~\ref{fig:tsne_vis_cub}, the pre-trained model aggregates birds with similar colors to the query regardless of class. However, after being trained by STML, the model only clusters birds of the same class.
Fig.~\ref{fig:tsne_vis_cars} shows that the embedding space of our model places the samples of the same class close to each other despite different colors and viewpoint changes, unlike the embedding space of the pre-trained model where unrelated samples are aggregated. In Fig.~\ref{fig:tsne_vis_sop}, the pre-trained model clusters samples of similar categories, whereas our model successfully gathers samples belonging to some parts of the query image.

\begin{figure*} [!t]
\centering
\includegraphics[width = 0.85 \textwidth]{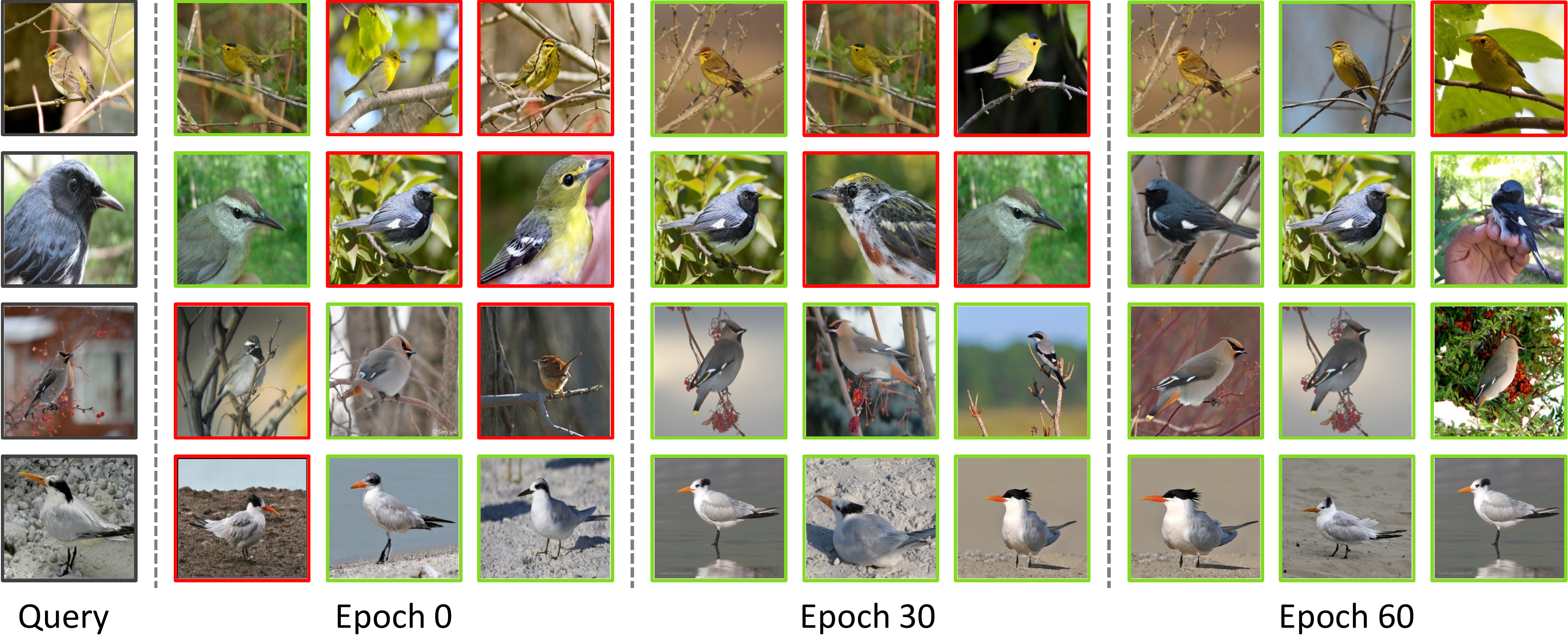}
\caption{
Top-3 retrievals of our model at every 30 epochs on the CUB-200-2011 datasets. Images with green boundary are correct and those with red boundary are incorrect.
} 
\label{fig:quals_cub}
\end{figure*}

\begin{figure*} [!t]
\centering
\includegraphics[width = 0.85 \textwidth]{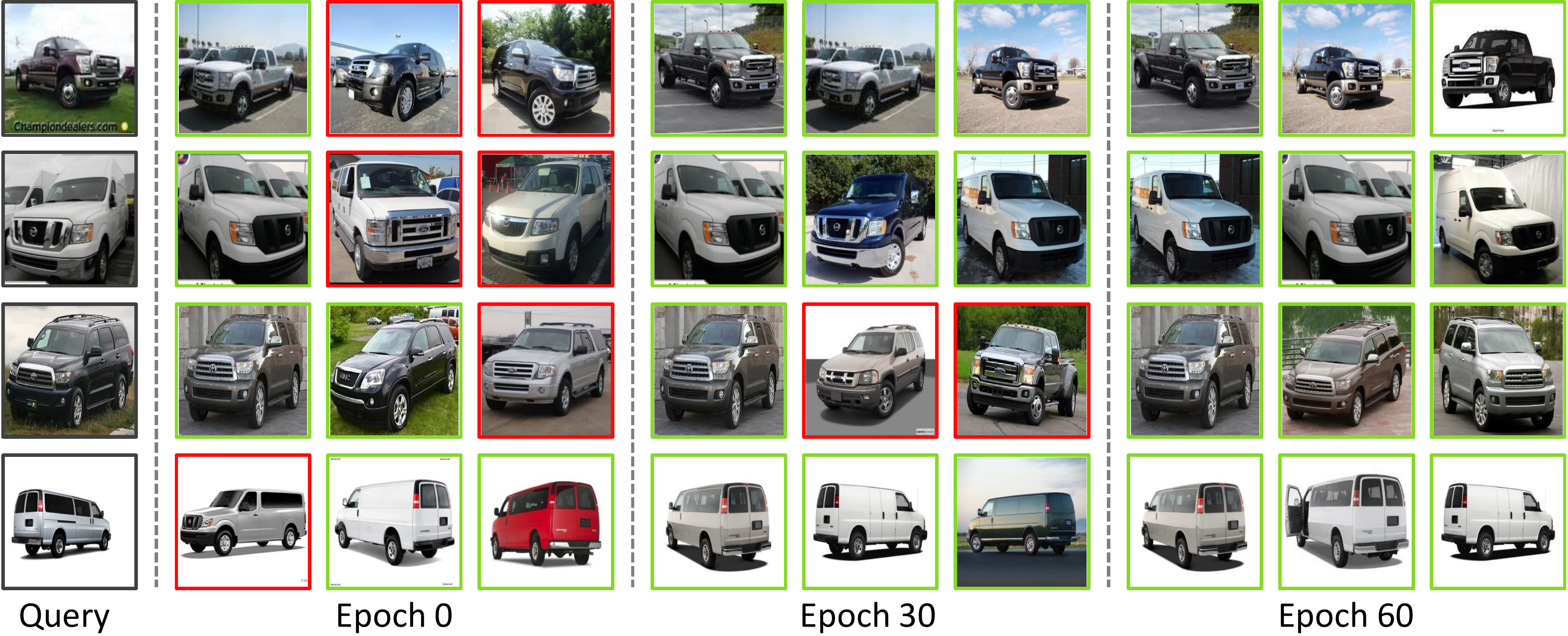}
\caption{
Top-3 retrievals of our model at every 30 epochs on the Cars-196 datasets. Images with green boundary are correct and those with red boundary are incorrect.
} 
\label{fig:quals_cars}
\end{figure*}

\begin{figure*} [!t]
\centering
\includegraphics[width = 0.85 \textwidth]{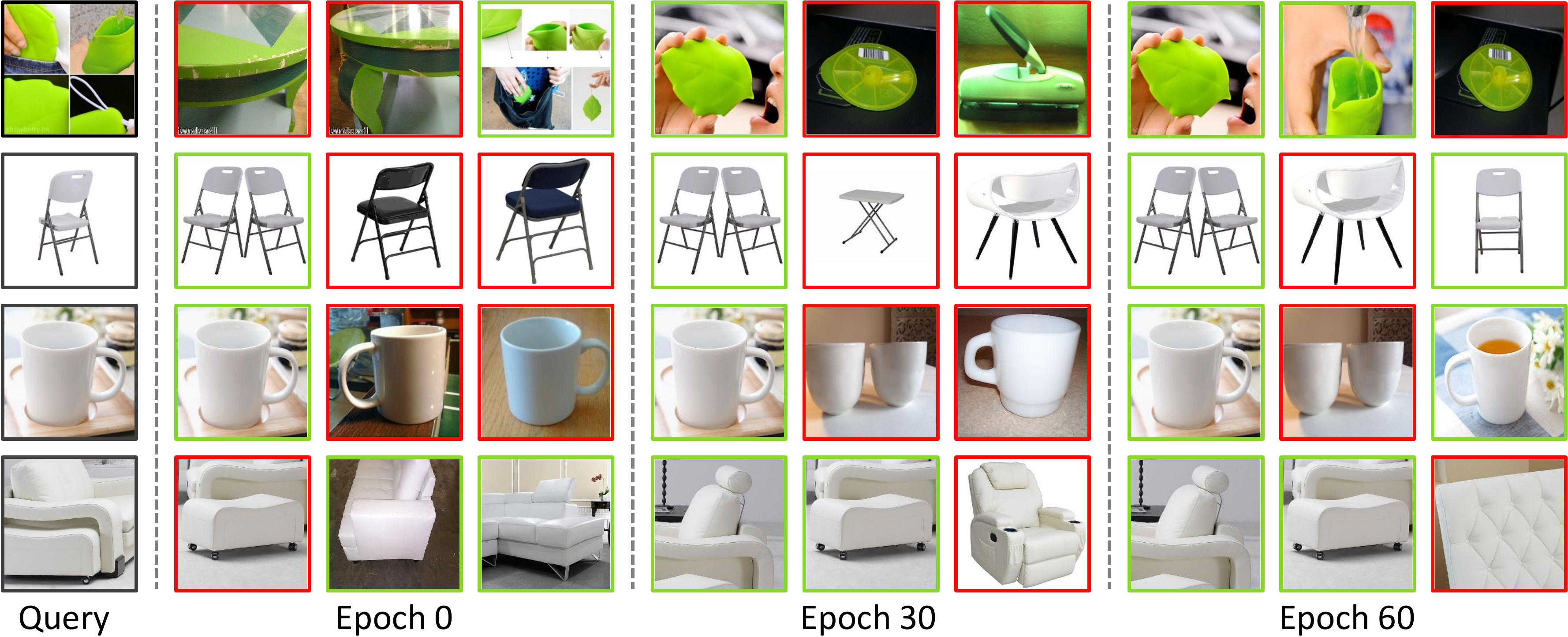}
\caption{
Top-3 retrievals of our model at every 30 epochs on the SOP datasets. Images with green boundary are correct and those with red boundary are incorrect.
} 
\label{fig:quals_SOP}
\end{figure*}

In addition, more qualitative retrieval results of our model at every 30 epochs of training on the three datasets are presented in Fig.~\ref{fig:quals_cub}, \ref{fig:quals_cars}, and \ref{fig:quals_SOP}. As shown in the figures, at the beginning of training, the model tends to retrieve only samples that are similar in appearance to the query, without understanding their semantic relations. As training progresses, the model trained by STML distinguishes samples of the same class more accurately among visually similar samples. 

All the results in these figures are obtained from unseen class samples without any manual annotation. 
The results suggest that contextualized semantic similarity estimates the class-equivalence relations between samples quite accurately, and thus STML allows the final embedding model to generalize well even to unseen classes through reliable pseudo labels considering semantic relations between data.

\end{document}